\documentclass[orivec]{imports/llncs}
\usepackage[utf8]{inputenc}
\usepackage{amssymb}
\usepackage{amsmath}
\usepackage{amsbsy}
\usepackage[ruled,noend,linesnumbered]{algorithm2e}
\usepackage{tikz}
\usepackage{changepage}
\usetikzlibrary{positioning}
\usetikzlibrary{arrows}
\usepackage{hyperref}
\usepackage{mdsymbol}
\usepackage{caption}
\usepackage{subcaption}
\usepackage{mathtools}
\usepackage{array, boldline, makecell, booktabs}

\newenvironment{algo}
 {\par\addvspace{\topsep}
  \centering
  \begin{minipage}{\linewidth}
  \hrule\kern2pt}
 {\par\kern2pt\hrule
  \end{minipage}
  \par\addvspace{\topsep}}

\begin{document}

\title{Inferring Capabilities by Experimentation}
\institute{Robotics Institute, Carnegie Mellon University \and Department of Machine Learning, Carnegie Mellon University}
\author{Ashwin Khadke\inst{1} \and Manuela Veloso\inst{2}}
\maketitle

\begin{abstract}
We present an approach to enable an autonomous agent (learner) in building a model of a new unknown robot's (subject) performance at a task through experimentation. The subject's appearance can provide cues to its physical as well as cognitive capabilities. Building on these cues, our active experimentation approach learns a model that captures the effect of relevant extrinsic factors on the subject's ability to perform a task. As personal robots become increasingly multi-functional and adaptive, such autonomous agents would find use as tools for humans in determining "What can this robot do?". We applied our algorithm in modelling a NAO and a Pepper robot at two different tasks. 
We first demonstrate the advantages of our active experimentation approach, then we show the utility of such models in identifying scenarios a robot is well suited for, in performing a task.
\end{abstract}
\section{Introduction}


Suppose we get a multipurpose domestic robot for our home but are not familiar with all of its functionalities. How do we identify what tasks it can perform? Appearance and specifications of a robot can convey information about its physical as well as cognitive capabilities. Seeing a legged robot equipped with a camera and a microphone could make one wonder if it can climb stairs, recognize faces, detect hand gestures or interpret voice commands. How do we identify which of these appearance-deduced tasks it can actually perform? Moreover, although physical appearances can provide rich cues about a robot's capabilities they are not sufficient to identify the scenarios in which it can function well. The robots Roomba and Braava appear similar and are both used for cleaning floors. But the Roomba can clean carpets and not wet floors, which is exactly the opposite of what Braava can do. For a human working collaboratively with a robot, knowing the robot's strengths and shortcomings is especially useful. 
 A robot's spec sheet provides information about different sensors and actuators. But inevitably how well a robot performs a task depends on the way it is programmed and for a naive user, this is difficult to determine simply based on appearance and specifications.
Experimenting with a robot can help identify the scenarios it is well suited for. But experimentation is tedious and intelligent robots are capable of learning new skills and adapting to new scenarios over time. This motivates the need for an autonomous system that can intelligently experiment with robots, identify their skills and quantify their applicability in different scenarios. 


In this paper we tackle the problem of an autonomous agent (learner) building a model of a robot (subject) at performing a particular task through experimentation. The outcomes of these experiments can be non-deterministic. We call such models as Capability Models (Section \ref{sec:ABM}). 
We assume the learner can infer tasks a subject can potentially perform from its appearance and, present a method to build a model of the subject at one of these tasks. 
Apart from the subject's inherent capabilities, certain extrinsic factors may affect its performance at the task. Assuming some of these factors are controllable and the learner can choose values for such factors in the experiments it conducts, we provide an approach to pick values for controllable factors that generate the most informative outcomes (Section \ref{subsec:albn}). However, knowing the set of extrinsic factors relevant for a particular robot and a task a priori is not always feasible. We present a model refinement method to identify relevant factors from a set of candidates (Section \ref{subsec:mrefine}). Further we show that Capability Models can be used in quantifying a robot's ability to perform a task in different scenarios thereby identifying situations a robot is well suited for (Section \ref{sec:capab}). 
We applied our algorithm in modelling a NAO robot at the task of kicking a ball. We show that active experimentation leads to faster learning of the subject's model than passively observing it perform and that our model refinement approach correctly identifies the set of relevant factors missing from the model (Section \ref{subsec:ballkick}). Furthermore, we learned a Capability Model for a Pepper robot programmed to pick up and clear objects off a table. From the learned model, we identified the types of objects the robot can pick up. We demonstrate how this knowledge is useful in improving performance at a collaborative clear-the-table task (Section \ref{subsec:mapplicable}).



\vspace{-0.1cm}
\section{Related Work}
Earlier works on learning from experimentation either addressed domains that are inherently deterministic \cite{Mitchell:1993,Scott:1993} or assumed that the experiments they conduct are deterministic \cite{Gil:1994}. 
Owing to noisy actuation and sensing, outcomes of experiments with robotic systems are non-deterministic and, hence the problem of deducing a robot’s capabilities through experimentation is challenging. 

Affordance \cite{Gibson:1979} is a relation between a certain effect, a class of objects and certain robot action. Learning affordances \cite{Sahin:2007,Dearden:2005} is similar to learning the physical capabilities of a robot. However, these approaches learn a mapping from motor commands to effects on different objects characterized by raw sensory input. It is difficult for a human to use these models in identifying scenarios a robot is well suited for.
Another approach \cite{Montesano:2008} uses visual features (Size, Shape, Color etc.) to characterize objects and learns models that capture the effect (Object speed etc.) of a robot's actions (Tap, Push, etc.) on different objects. However, all of these works \cite{Sahin:2007,Dearden:2005,Montesano:2008} adopt passive approaches to learn and do not use their existing models to reason about what to explore next. Moreover, they provide no concrete method to quantify a robot's capability.

Modeling a robot's capability can be thought of as learning a forward model \cite{Jordan:1992}. Such models predict the change in a robot's state
brought about by its actions.
Active approaches to learn forward models exist \cite{Baranes:2009,Baranes:2013,Wang:2014,Forestier:2016,Forestier:2017a}. However, these models 
only predict the immediate effects of an action i.e. predict the next state given the current state and action and provide no approach to identify from the forward model, the likelihood of successful task execution in different scenarios. An idea to mitigate this problem is to learn a forward model over higher level states \cite{Hofer:2016,Mugan:2012}. But the emphasis in these works is on learning a policy for a task \cite{Wang:2014,Forestier:2016} or a combination of tasks \cite{Baranes:2013,Forestier:2017a} or learning higher level actions (sub-policies) relevant for a task \cite{Hofer:2016}.

\vspace{-0.2cm}
\section{Capability Models}\label{sec:ABM}
Suppose the subject is an anthropomorphic robot (Fig. \ref{fig:nao}) and the learner chooses to build a model of the subject at the task of kicking a ball. 
Extrinsic factors that the learner may consider include, the size of the ball and turf on which the subject is playing.
Among these factors, ball size is controllable. 
An experiment for this task would constitute the learner commanding the subject to kick a ball in certain direction from a particular position and observing the outcome.
A robot's perception and actuation is noisy and therefore the outcome is not deterministic. To capture this non-determinism we use a Bayesian Network.

\vspace{-0.3cm}
\begin{figure}[!h]
\centering
\includegraphics[width=6cm,height=4.5cm]{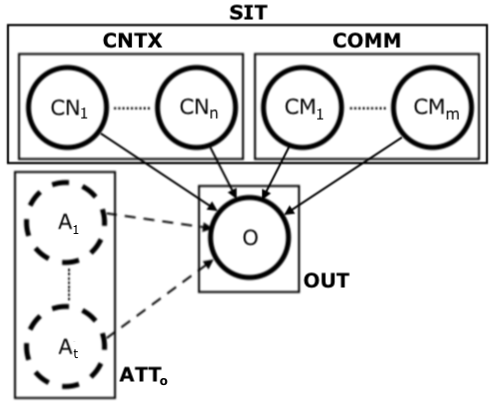}
\caption{Capability Model}
\label{fig:ABM}
\end{figure}
\vspace{-0.2cm}
We introduce Capability Model (Fig. \ref{fig:ABM}), a Bayesian Network which consists of three types of nodes namely:
\begin{itemize}
\item {
SIT = CNTX $\cup$ COMM, is the set of variables that describe the situation in which the subject is performing the task.
\begin{itemize}
\item CNTX is the set of extrinsic factors (context) for the task.
\item COMM is the set of commands given to the subject.
\end{itemize}
}
\item OUT is the set of variables denoting the outcomes of the task. 
\item $\text{ATT}_o$ is the set of attributes of variable $o\in\text{OUT}$. These variables are not explicitly accounted in the model. Section \ref{subsec:mrefine} discusses their need and utility. 
\end{itemize}
We capture the subject's ability to perform a task in the conditional probability tables associated with this Bayesian Network.

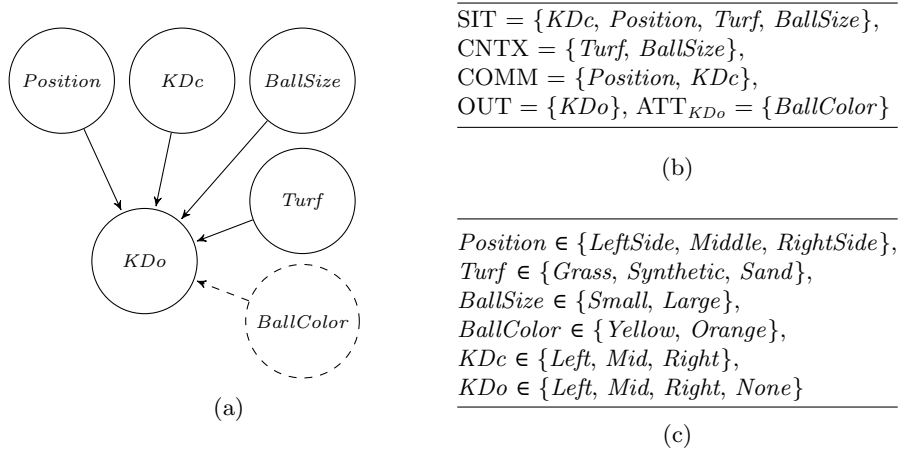
\begin{figure}
\begin{subfigure}{.48\textwidth}
\begin{tikzpicture}
[->,>=stealth',shorten >=1pt,node distance=1.6cm, minimum size=1.4cm, main node/.style={circle,draw},additional node/.style={circle,dashed,draw}]
  \node[main node] (KDc) {\scriptsize{\textit{KDc}}};
  \node[main node] (KP) [left of=KDc] {\scriptsize{\textit{Position}}};
  \node[main node] (BallSize) [right of=KDc] {\scriptsize{\textit{BallSize}}};
  \node[main node] (KDo) [below right=1.4cm and 0.1cm of KP] {\scriptsize{\textit{KDo}}};
  \node[main node] (Turf) [below of=BallSize] {\scriptsize{\textit{Turf}}};
  \node[additional node] (BallColor) [below of=Turf] {\scriptsize{\textit{BallColor}}};
  \path[every node/.style={font=\sffamily\small}]
    (KDc) edge [right] node[left] {} (KDo)
    (KP) edge node [right] {} (KDo)
    (BallSize) edge [right] node {} (KDo)
    (BallColor) edge [dashed] node {} (KDo)
    (Turf) edge [right] node {} (KDo);
\end{tikzpicture}
\caption{}
\end{subfigure}  
\begin{subfigure}{.48\textwidth}
    \null
    \begin{algo}
    SIT = \{\textit{KDc}, \textit{Position}, \textit{Turf}, \textit{BallSize}\},\\
    CNTX = \{\textit{Turf}, \textit{BallSize}\}, \\
    COMM = \{\textit{Position}, \textit{KDc}\},\\ OUT = \{\textit{KDo}\},
    ATT$_{\text{\textit{KDo}}}$ = \{\textit{BallColor}\}
\vspace{-0.2cm}

\noindent\hrulefill 
\vspace{0.15cm}
\caption{}
\label{mballkickset}
\vspace{0.2cm}
\noindent\hrulefill 

\vspace{0.02cm}
   \textit{Position} $\in$ \{\textit{LeftSide}, \textit{Middle}, \textit{RightSide}\},\\
    \textit{Turf} $\in$ \{\textit{Grass}, \textit{Synthetic}, \textit{Sand}\},\\
    \textit{BallSize} $\in$ \{\textit{Small}, \textit{Large}\},\\
    \textit{BallColor} $\in$ \{\textit{Yellow}, \textit{Orange}\},\\
    \textit{KDc} $\in$ \{\textit{Left}, \textit{Mid}, \textit{Right}\},\\
    \textit{KDo} $\in$ \{\textit{Left}, \textit{Mid}, \textit{Right}, \textit{None}\}
 \end{algo}
 \vspace{-0.3cm}
\caption{}
\label{mballkickvars}
\end{subfigure}
\vspace{-0.1cm}
\caption{Capability Model for the \textit{BallKick} task. (a) depicts the Bayesian Network, (b) shows the type of each variable in the model and, (c) describes the values each variable can take.}
\label{mballkick}
\end{figure}

 Fig. \ref{mballkick} presents a Capability Model for the \textit{BallKick} task discussed before using the notation we just introduced.
 \textit{Position} represents the location of the ball with respect to the subject before it kicks. 
\textit{KDc} and \textit{KDo} denote the commanded and observed  kick direction respectively. \textit{KDo} is \textit{None} if the subject attempts but fails to kick the ball or does not attempt to kick. 
The set OUT need not always be a singleton. How far a subject kicks a ball could be another outcome for the \textit{BallKick} task. 
\vspace{-0.3cm}

\vspace{-0.85cm}
\section{Building Capability Models} \label{sec:labm}

Building a model involves identifying the right factors to include in the Bayesian Network and learning the conditional probabilities associated with it. First, we present a method to learn the conditional probabilities assuming the learner knows the right factors and structure of the network is fixed. Later, we describe our method to refine the model if need be. Here we introduce some notation.

\begin{itemize}
\item{A Bayesian Network is a tuple $(G,\theta)$
    \begin{itemize}
        \item $G\equiv(V,E)$ is a graph, $V$ is the set of nodes and $E$ is the set of edges.
        \item $V = \text{SIT} \cup \text{OUT}$ are random variables with Multinomial distribution
        \item $E$ capture the conditional dependencies amongst the nodes $V$
        \item $\theta$ parameterize the conditional probability distributions
    \end{itemize}
}
\vspace{0.1cm}
\item $\text{Q}\subset\text{SIT}$ are variables a learner can control in an experiment. COMM $\subset$ Q. An instantiation\footnotemark of Q is called a \textit{Query}.
\vspace{0.1cm}
\item Instantiations
of CNTX, COMM
and OUT are called \textit{Context}, \textit{Command}
and \textit{Outcome} respectively. $\text{\textit{Situation}}=\text{\textit{Context}}\cup\text{\textit{Command}}$\label{situationnotation}
\end{itemize}
\footnotetext{\label{instanfootnote}An instantiation of a set of random variables is a mapping from variables in the set to values in their domain. \textit{Query} $\gets \cup_{q\in \text{Q}}\{q:v_{\text{\textit{q}}}\}$ where $v_{\text{\textit{q}}}\in \text{\textit{Domain(q)}}$
}
\subsection{Active Learning for Bayesian Networks} \label{subsec:albn}
We adopt a Bayesian approach to learn parameters and use the algorithm presented in \cite{Tong:2000} to build a distribution over $\theta$. We describe it here in short.

The algorithm starts with a prior $p(\theta)$ and builds a posterior $p'(\theta)$ by actively experimenting with the subject. In each experiment, it picks a \textit{Query} and requests the subject to perform the task. Variables in $\text{SIT}\setminus\text{Q}$ either have a fixed value (Eg. \textit{Turf} in the \textit{BallKick} task) or are assigned some value by the environment. A standard Bayesian update on the prior $p$, for the parameters of the conditional distributions identified by the \textit{Situation} i.e. $P(o|\text{\textit{Situation}})\hspace{0.1cm}\forall o\in\text{OUT}$, based on the \textit{Outcome} yields the posterior $p'$. The posterior $p'$ becomes the prior for the next experiment.


To generate a \textit{Query} from the current estimate of $p(\theta)$ we need a metric to evaluate how good the current estimate is. We can then quantify the improvement in the estimate brought about by different queries and pick the one which leads to the biggest improvement. Let $\theta^\star$ be the true parameters of the model and $\theta'$ be a point estimate. $\sum_{o\in\text{OUT}}D_{\text{\textit{KL}}}(P_{\theta^\star}(o) || P_{\theta'}(o))$\footnote{\label{kldfootnote} $D_{\text{\textit{KL}}}(P_1(o)||P_2(o)) = \smashoperator{\sum_{v_o\in\text{\textit{Domain}}(o)}} P_1(o=v_o)\text{log}\Big(\frac{P_1(o=v_o)}{P_2(o=v_o)}\Big)$, $P_1$ and $P_2$ are distributions of $o$.} denotes the error in point estimate $\theta'$, where $P_{\theta^{\star}}(o)$ and $P_{\theta'}(o)$ are distributions of variable $o$ parameterized with $\theta^{\star}$ and $\theta'$ respectively.
$\theta^{\star}$ is not known, but we do have $p(\theta)$, which is our belief of what $\theta^\star$ is given the prior and observations.  
Error in point estimate $\theta'$ with respect to $p(\theta)$ can be quantified as in Eq. (\ref{riskt}) 
\begin{eqnarray}
&\label{riskt}
\text{Error}_{p}(\theta') = \sum_{o\in\text{OUT}}\int_{\theta}D_{\text{\textit{KL}}}(P_{\theta}(o) || P_{\theta'}(o))p(\theta)d\theta \\
&\label{riskp}
\text{ModelError}(p) = \displaystyle\min_{\theta'}\text{Error}_{p}(\theta')
\end{eqnarray}
We use ModelError$(p)$ of a distribution $p$, defined in Eq. (\ref{riskp}), as the measure of quality for the estimate $p(\theta)$. Lower the ModelError, better the estimate. 
We would want to see observations that reduce the ModelError associated with $p(\theta)$ to improve our estimate. But the learner can only control the \textit{Query}.
For a particular \textit{Query} we take an expectation over possible observations to evaluate the Expected Posterior Error (EPE) as defined in Eq. (\ref{epr}). In every experiment the learner picks the \textit{Query} with the lowest EPE.
\begin{equation}
\label{epr}
\text{EPE}(p,\text{\textit{Query}}) = E_{\Theta\sim p(\theta)}\big(E_{\textit{Outcome}\sim P_{\Theta}(\text{OUT}|\text{\textit{Query}})}\big(\text{ModelError}(p')\big)\big)
\end{equation}
In Eq. (\ref{epr}) $p'$ represents the posterior obtained after updating prior $p$ with sample drawn from $P_{\Theta}(\text{OUT}|\text{\textit{Query}})$. For further details, please refer \cite{Tong:2000}. $\theta_T$ as defined in Eq. (\ref{eq:thetalearned}) parameterize the conditional probabilities of the Capability Model for task $T$. We compute $\theta_T$ using the learned distribution $p(\theta)$.  
\begin{equation}\label{eq:thetalearned}
    \theta_{T} = \int_{\theta}\theta p(\theta)d\theta
\end{equation}
\vspace{-0.2cm}

\subsection{Model Refinement} \label{subsec:mrefine}
Every subject may have a different set of extrinsic factors relevant for a task.
If a subject only detects balls of a certain color, a variable \textit{BallColor} should be included in the model for the \textit{BallKick} task. We assume the learner starts with a minimal set CNTX and variables representing some of the relevant factors may be missing. 
In Section \ref{sec:ABM} we defined ATT$_o$ to be attributes of variable $o\in\text{OUT}$ not explicitly accounted in the model. We assume that the missing variables, if any, belong to these sets. 
Including all of them makes the model unnecessarily large and difficult to learn. We need a metric to quantify the dependence of variables in $\text{OUT}$ on the attributes to identify the relevant ones. 


An attribute $\text{A}_{\text{j}}\in\text{ATT}_o$ is relevant if the subject's performance (distribution of $o$) is drastically different for different values of $\text{A}_{\text{j}}$ in at least one of the observed \textit{Situation}s.  
In each experiment, attributes are chosen randomly independent of the \textit{Situation}. 
After every experiment, the learner computes $\hat{P}(\text{A}_{\text{j}}|\textit{Situation})$ and $\hat{P}(o|\text{A}_{\text{j}},\textit{Situation}) \hspace{0.05cm}\forall o\in\text{OUT}, \forall \text{A}_{\text{j}}\in\text{ATT}_o$. $\hat{P}$ is the estimate of the true distribution from the observations in the past experiments. 
We use the metric defined in Eq. (\ref{coeffmutualInf}), to quantify the dependence of $o$ on the attribute $\text{A}_{\text{j}}$ in a \textit{Situation}.
\vspace{-0.1cm}
\begin{equation}
\label{coeffmutualInf}
\text{R}(o,\text{A}_{\text{j}}|\textit{Situation}) = \frac{\text{I}(o,\text{A}_{\text{j}}|\textit{Situation})}{\min(H(\hat{P}(o|\textit{Situation})),H(\hat{P}(\text{A}_{\text{j}}))}
\end{equation}
\vspace{-0.1cm}
\begin{equation}
 \begin{split}
\label{mutualInf}
 \text{I}(o,\text{A}_{\text{j}}|\textit{Situation}) = H(\hat{P}(o|\textit{Situation})) -\\ & \hspace{-1cm}\displaystyle\smashoperator{\sum_{a\in \text{\textit{Domain}}_{\text{\textit{valid}}}(\text{A}_{\text{j}}|\textit{Situation})}}
 H(\hat{P}(o|\text{A}_{\text{j}}=a,\textit{Situation}))\hat{P}(\text{A}_{\text{j}}=a) \hspace{0.25cm}
 \end{split}
\end{equation}
In Eq. (\ref{mutualInf}), $H(P)$ is the entropy of distribution $P$ and $\text{\textit{Domain}}_{\text{\textit{valid}}}(\text{A}_{\text{j}}|\textit{Situation})$ are values of $\text{A}_{\text{j}}$ that have been observed sufficiently in a \textit{Situation}. 
As $\text{A}_{\text{j}}$ is sampled independently of the \textit{Situation}, $\hat{P}(\text{A}_{\text{j}})\approx\hat{P}(\text{A}_{\text{j}}|\textit{Situation})$. Therefore, $\text{I}(o,\text{A}_{\text{j}}|\textit{Situation})$ is approximately the Mutual Information of $o$ and $\text{A}_{\text{j}}$ given the \textit{Situation}, and $\text{R}(o,\text{A}_{\text{j}}|\textit{Situation})$ is the Coefficient of Mutual Information. 
If $\text{R}(o,\text{A}_{\text{j}}|\textit{Situation})$ is greater than threshold $\text{R}_{\text{Th}}$, the learner adds $\text{A}_{\text{j}}$ to CNTX, updates the parents of variable $o$ in the graph and creates conditional probability tables for the updated graph.

Algorithm 
\ref{alg:modelCorr}
outlines the overall method. 
Function BestQuery (Line \ref{line:bestquery}) computes a \textit{Query} using the approach described in Section \ref{subsec:albn}. 
Function  IdentifyDependence (Line \ref{line:idendep}) applies the model refinement method presented above.
\begin{algorithm}
\SetAlgoLined
\caption{\textbf{LearnModel}(maxIter, $\text{R}_{\text{Th}}$)}
\label{alg:modelCorr}
 From Domain Knowledge
 
    \quad Construct $G\equiv(V,E)$
    
    \quad Initialize Q, $p(\theta)$
    
SituationsObserved $\gets \emptyset$

Initialize $\hat{P}$
    
\textit{i} $\gets 0$ 
    
\While{i $<$ \text{\upshape maxIter}}{
	$i\gets i+1$
	
    ATTIncluded $\gets$ \{\}
    
	\textit{Query} $\gets$ \textbf{BestQuery}($p(\theta)$, Q) \label{line:bestquery}
	
    \textit{Attributes} $\gets$ \textbf{SampleUniform}\big($\bigcup_{o\in\text{OUT}}\text{ATT}_o$\big)
    
	\textit{Outcome}, \textit{Situation} $\gets$ \textbf{Experiment}(\textit{Query}, \textit{Attributes})
	
    SituationsObserved $\gets$ SituationsObserved $\bigcup$ \textit{Situation}
    
	$p(\theta), \hat{P} \gets \text{\textbf{Update}}(p(\theta), \hat{P}, \text{\textit{Situation}}, \text{\textit{Outcome}}, \text{\textit{Attributes}})$
	
    \For{$o\hspace{0.05cm}\in\hspace{0.05cm}\text{\upshape OUT}$}{
    	$\text{ATTIncluded}[o] \gets \text{\textbf{IdentifyDependence}}(\hat{P}, \text{ATT}_o, \text{SituationsObserved}, \text{R}_{\text{Th}})$} \label{line:idendep}
    $p(\theta), G, \text{Q} \gets \text{\textbf{Modify}}(p(\theta), G, \text{Q}, \text{ATTIncluded})$}
\Return{$(G,p(\theta))$}
\end{algorithm}
\vspace{-0.55cm}
\section{Quantifying Capabilities} \label{sec:capab} 
 To determine how well a robot performs a task in different scenarios, we need a reference that captures the expected performance and, a metric that quantifies how the robot fares against this standard. We assume the reference for task $T$, is a distribution of the outcome variables conditioned on the commands i.e. $P^{T}_{\text{ref}}(\text{OUT}|\text{COMM})$. Eq. (\ref{eq:pexpect}) shows a possible reference for  \textit{BallKick} task (Fig. \ref{mballkick}).
 \begin{equation}\label{eq:pexpect}
    P^{\text{\textit{BallKick}}}_{\text{ref}}(\text{\textit{KDo}}|\text{\textit{KDc}},\text{\textit{Position}})=
    \left\{
    \begin{array}{ll}
    1& \text{if \textit{KDo} = \textit{KDc}}\\
    0 & \text{otherwise} \\
    \end{array}
    \right.
 \end{equation}
 This reference implies that a robot is expected to always kick in the commanded direction. $\text{\textit{Score}}_{\text{\textit{T}}}(\text{\textit{Context}})$, defined in Eq. (\ref{eq:contextcapab}), denotes how well a robot fares at a task \textit{T} in certain \textit{Context}. Lower score values indicate poor performance. We say a robot functions well in a \textit{Context}, if the score is higher than a threshold.
\begin{equation}
     \text{\textit{Mismatch}}(\text{\textit{Context}})=\frac{\displaystyle\smashoperator{\sum_{\text{\textit{Command}}}}D_{\text{\textit{KL}}}\big(P_{\text{ref}}^T(\text{OUT}|\text{\textit{Command}})||P_{\theta_T}(\text{OUT}|\text{\textit{Situation}})\big)\footnotemark\textsuperscript{,}\footnotemark}{|\text{\textit{Domain}(COMM)}|}\hspace{0.4cm}
\end{equation}
\begin{equation}
   \text{\textit{Score}}_{\text{\textit{T}}}(\text{\textit{Context}})= \frac{1}{1+\text{\textit{Mismatch}}(\text{\textit{Context}})
     }\label{eq:contextcapab}
\end{equation}
\addtocounter{footnote}{-1}
\footnotetext{$P_{\theta_T}$ is the distribution parameterized by $\theta_T$. For a task $T$, $\theta_T$ is defined in Eq. (\ref{eq:thetalearned})}

\vspace{-0.5cm}
\section{Results} \label{sec:experiments}
\begin{figure}
\begin{subfigure}[b]{0.45\linewidth}
\centering
\includegraphics[height=6cm,width=4cm]{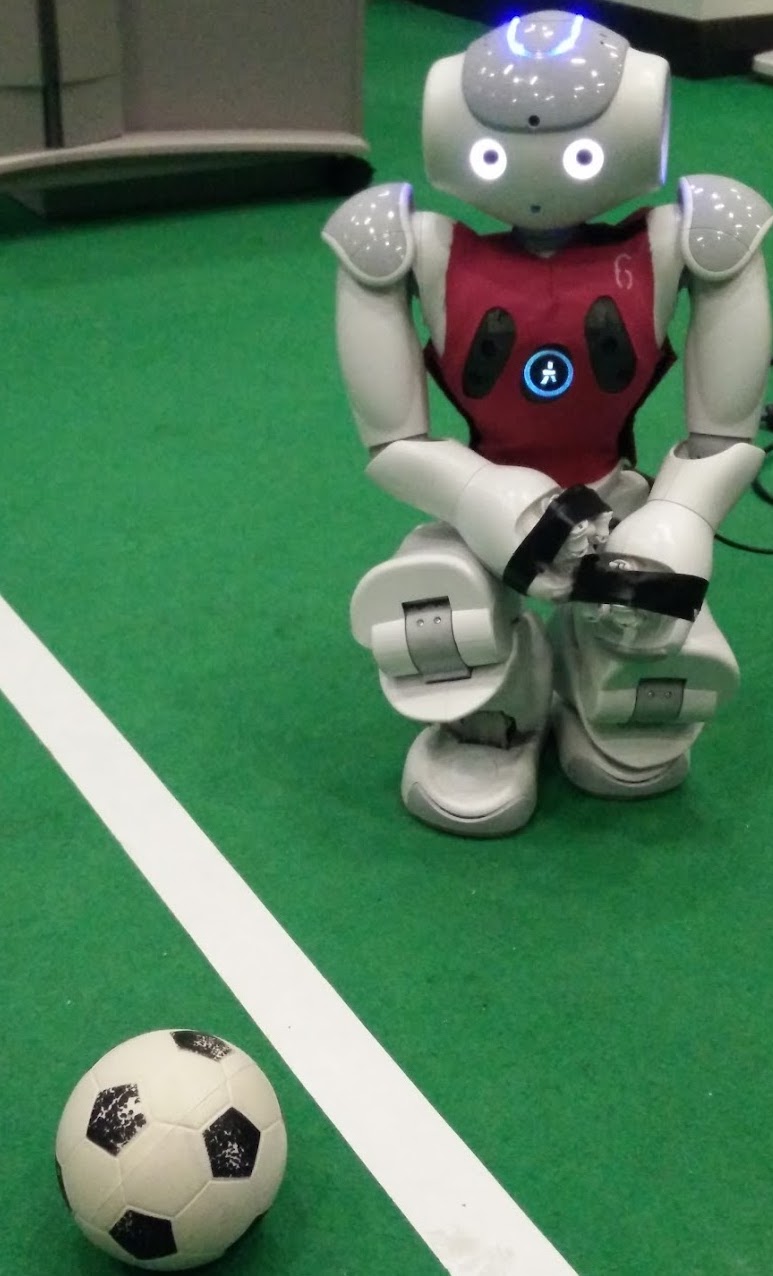}
\caption{NAO} 
\label{fig:nao}
\end{subfigure}
\hfill
\begin{subfigure}[b]{0.45\linewidth}
\centering
\includegraphics[height=6cm,width=5cm]{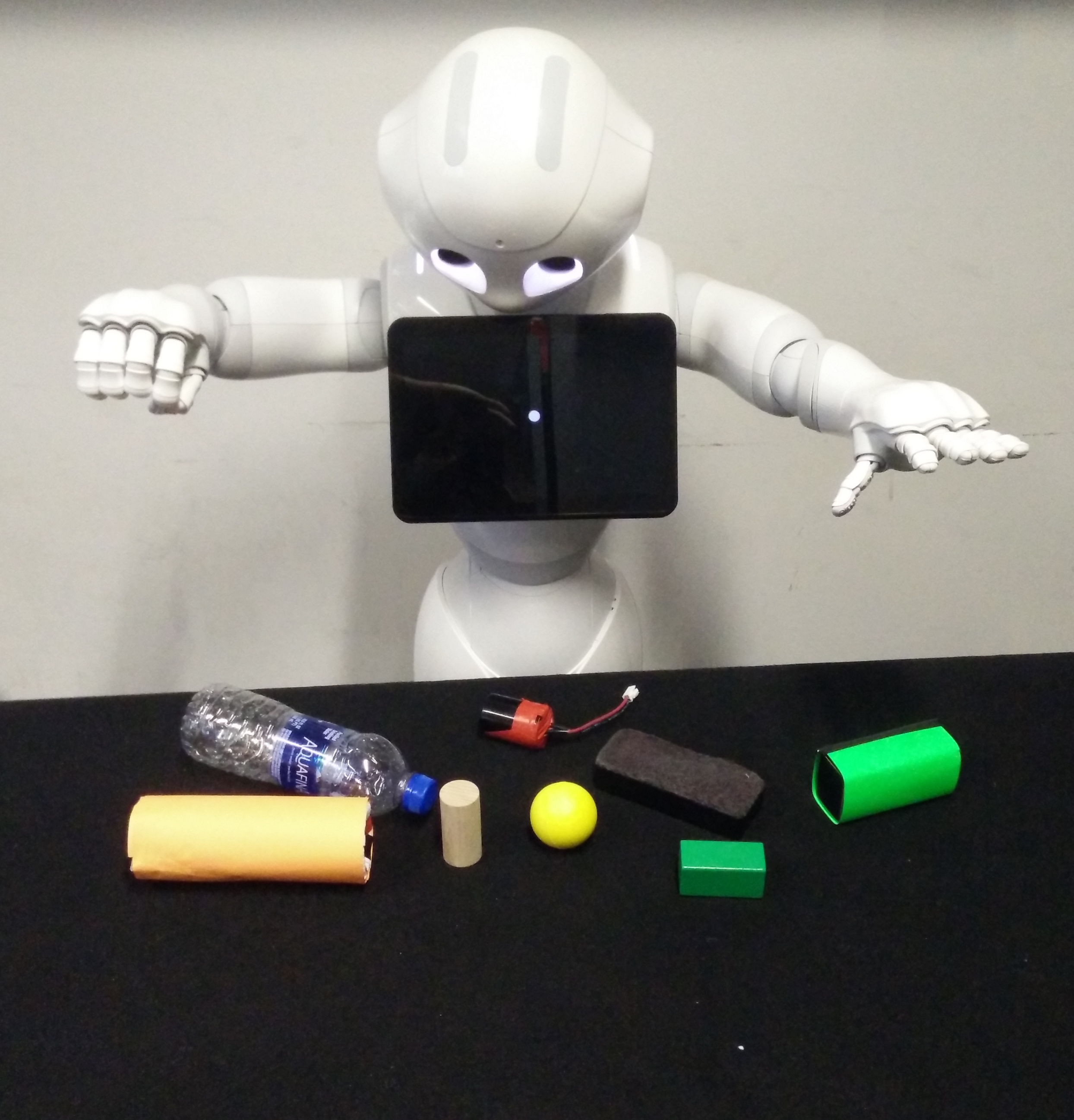}
\caption{Pepper}
\label{fig:pepper}
\end{subfigure}
\caption{Subjects for experiments}
\end{figure}

We present results of applying our method in modelling a NAO robot (Fig. \ref{fig:nao}) performing the task of kicking a ball and a Pepper robot (Fig. \ref{fig:pepper}) picking up different types of objects and clearing them off a table.  
\vspace{-0.2cm}
\subsection{BallKick Task}\label{subsec:ballkick}
Here we demonstrate advantages of the active experimentation approach and our model refinement algorithm. 
Our approach builds a model, but without knowing the ground truth we cannot determine how good the learned model is. So we programmed the robot to behave according to a predefined model with parameters $\theta_{\text{\textit{predefined}}}$. We use $D_{\text{\textit{KL}}}(P_{\theta_{\text{\textit{predefined}}}}(\text{OUT})\hspace{0.01cm}||\hspace{0.01cm}P_{\theta_{\text{\textit{BallKick}}}}(\text{OUT}))$
to determine how close the learned and predefined models are. 
$\theta_{\text{\textit{BallKick}}}$ is defined in Eq. (\ref{eq:thetalearned}), for the \textit{BallKick} task. 
Fig. \ref{fig:ABMGraphCorrection} shows the learner's initial guess of the Bayesian Network.
\vspace{-0.6cm}

\subsubsection{Passive vs Active} 

Passively observing a subject, where you do not control the scenarios in which you witness it perform a task, is equivalent to randomly picking \textit{Situation}s to build a model. We learned a model using our approach and another one by randomly picking \textit{Situation}s. Fig. \ref{fig:ABMGraphOriginal} depicts the Bayesian Network for the predefined model. We experimented with a single ball on a synthetic turf and thus variables \textit{Turf} and \textit{BallSize} 
were dropped in the predefined model. A subtle point to note, the experiments were noisy. The learner chose a particular \textit{Situation}, sampled a direction from $P_{\theta_{\text{\textit{predefined}}}}(\text{\textit{KDo}}|\text{\textit{Situation}})$ and commanded the robot to kick in this direction. However, owing to noisy perception and actuation, sometimes the robot kicked in directions it wasn't commanded to. Fig. \ref{fig:original} shows that despite noisy experiments the active approach converged faster.
\vspace{-0.3cm}
\subsubsection{Model Refinement} 
\begin{figure}[t]
\begin{subfigure}{\linewidth}
\centering
\begin{tikzpicture}
[->,>=stealth',shorten >=1pt,node distance=1.5cm, minimum size=1.3cm, main node/.style={circle,draw},additional node/.style={circle,dashed,draw}]
  \node[main node] (KDc) {\scriptsize{\textit{KDc}}};
  \node[main node] (KP) [left of=KDc] {\scriptsize{\textit{Position}}};
  \node[additional node] (BallSize) [right of=KDc] {\scriptsize{\textit{BallSize}}};
  \node[main node] (KDo) [below right=1cm and 0.1cm of KDc] {\scriptsize{\textit{KDo}}};
  \node[additional node] (BallColor) [right of=BallSize] {\scriptsize{\textit{BallColor}}};
  \node[additional node] (Turf) [below of=BallColor] {\scriptsize{\textit{Turf}}};

  \path[every node/.style={font=\sffamily\small}]
    (KDc) edge [right] node[left] {} (KDo)
    (KP) edge node [right] {} (KDo)
    (BallSize) edge [dashed] node {} (KDo)
    (BallColor) edge [dashed] node {} (KDo)
    (Turf) edge [dashed] node {} (KDo);
\end{tikzpicture}
\caption{}
\label{fig:ABMGraphCorrection}
\end{subfigure}
\vspace{1em}
\begin{subfigure}[b]{0.45\linewidth}
\vspace{-0.4cm}
\centering
\begin{tikzpicture}
[->,>=stealth',shorten >=1pt,node distance=1.5cm, minimum size=1.3cm, main node/.style={circle,draw},additional node/.style={circle,dashed,draw}]
  \node[main node] (KDc) {\scriptsize{\textit{KDc}}};
  \node[main node] (KDo) [below = 0.3cm of KDc] {\scriptsize{\textit{KDo}}};
  \node[main node] (KP) [left of=KDc] {\scriptsize{\textit{Position}}};
  \path[every node/.style={font=\sffamily\small}]
    (KDc) edge [right] node[left] {} (KDo)
    (KP) edge node [right] {} (KDo);
\end{tikzpicture}
\caption{}
\label{fig:ABMGraphOriginal}
\end{subfigure}
\hfill
\begin{subfigure}[b]{0.45\linewidth}
\vspace{-0.4cm}
\centering
\begin{tikzpicture}
[->,>=stealth',shorten >=1pt,node distance=1.5cm, minimum size=1.3cm, main node/.style={circle,draw},additional node/.style={circle,dashed,draw}]
  \node[main node] (KDc) {\scriptsize{\textit{KDc}}};
  \node[main node] (KDo) [below =0.3cm of KDc] {\scriptsize{\textit{KDo}}};
  \node[main node] (KP) [left of=KDc] {\scriptsize{\textit{Position}}};
  \node[main node] (BallSize) [right of=KDc] {\scriptsize{\textit{BallSize}}};
  \path[every node/.style={font=\sffamily\small}]
    (KDc) edge [right] node[left] {} (KDo)
    (KP) edge node [right] {} (KDo)
    (BallSize) edge [right] node {} (KDo);
\end{tikzpicture}
\caption{}
\label{fig:ABMGraphExpanded}
\end{subfigure}
\vspace{-0.4cm}
\caption{(a) depicts the learner's initial guess of the model and, (b) and (c) depict the Bayesian Network of the predefined model. 
Nodes are as defined in
Fig. \ref{mballkickvars}
}
\end{figure}
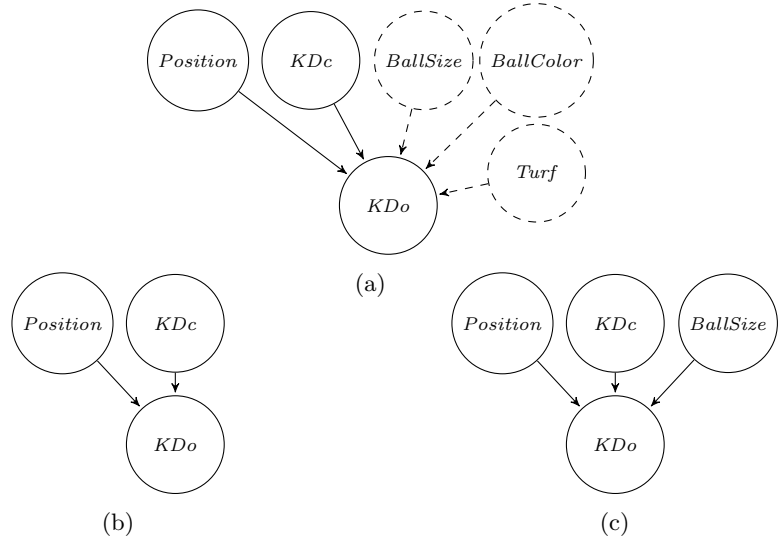

\begin{figure}[t]
\begin{subfigure}[b]{0.49\linewidth}
\centering
\includegraphics[width=6.0cm,height=4.0cm]{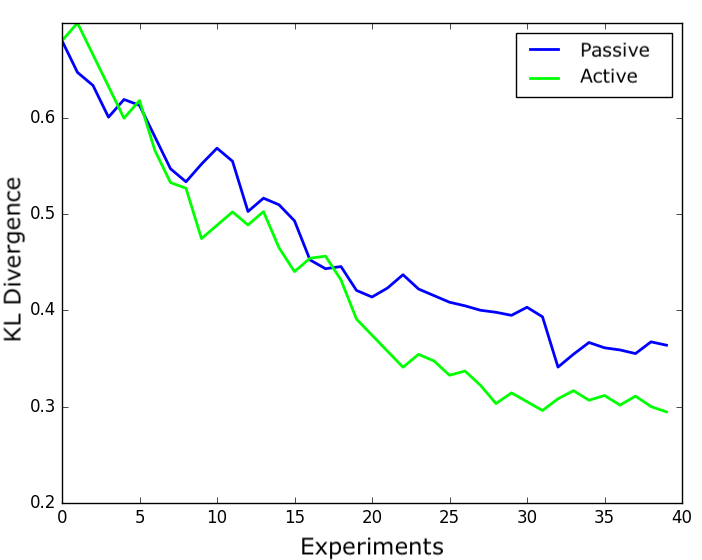}
\caption{} 
\label{fig:original}
\end{subfigure}
\hfill
\begin{subfigure}[b]{0.49\linewidth}
\centering
\includegraphics[width=6.0cm,height=4.0cm]{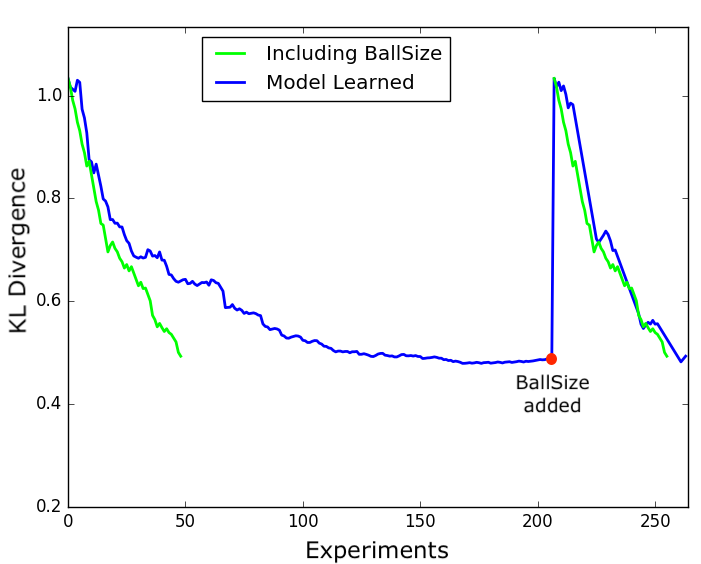}
\caption{}
\label{fig:correction}
\end{subfigure}
\vspace{1em}
\begin{subfigure}{\linewidth}
\centering
\includegraphics[width=6.0cm,height=4.0cm]{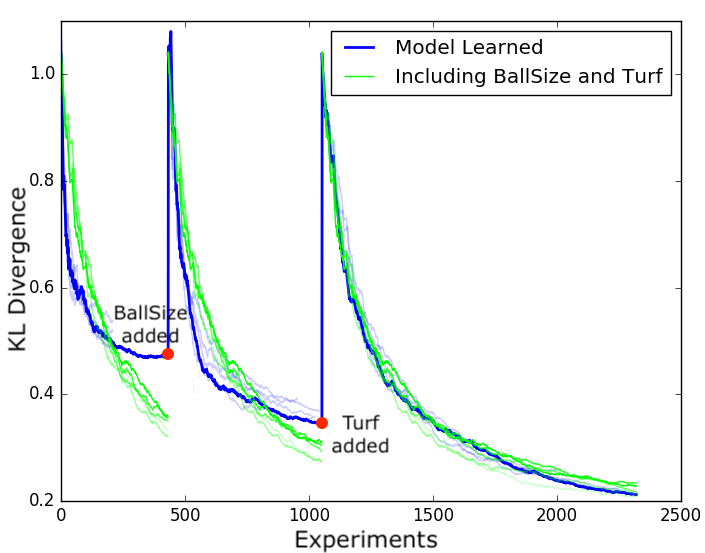}
\caption{}
\label{fig:extrasim}
\end{subfigure}
\vspace{-0.4cm}
\caption{(a), (b) and (c) depict the trend in $D_{\text{\textit{KL}}}(P_{\theta_{\text{\textit{predefined}}}}(\text{\textit{KDo}})\hspace{0.01cm}||\hspace{0.01cm}P_{\theta_{\text{\textit{BallKick}}}}(\text{\textit{KDo}}))$.
(a) compares the trends when learned actively vs passively. In (b) and (c), curves in blue depict trends for the learner's model as it identifies relevant factors to include, red points mark the instances when a new variable is added and curves in green depict the trend if the model were initialized with the right variables.
}
\end{figure}

We programmed the robot to detect balls of any color but only a specific size. If presented with a ball of a different size, the subject would not detect and thus won't kick i.e. \textit{KDo} would be \textit{None}. Fig. \ref{fig:ABMGraphExpanded} depicts the Bayesian Network for the predefined model and Fig. \ref{fig:ABMGraphCorrection} shows the learner's initial guess. 
The learning curve for this model is shown in blue in Fig. \ref{fig:correction}. The learner correctly identified the missing variable to be \textit{BallSize}. Once identified, the algorithm resets the conditional probabilities (hence the jump in the trend) and restarts the learning process with the updated model. To show that incorporating relevant attributes yields a model better representative of the subject, we learned another model including \textit{BallSize} from the start. As can be seen in Fig. \ref{fig:correction}, the model that included \textit{BallSize} from the start (in green) converged to a lower \textit{KL Divergence} compared to the model that did not (in blue). Moreover, post refinement the trends for both models are almost same. 

We tested in simulation, how our approach fares as the number of missing attributes increase. In these experiments the robot could only detect balls of a certain size and the kicked ball would randomly end up in any of the three possible directions
i.e. \textit{BallSize} and \textit{Turf} were the missing relevant variables. To simulate noisy experiments we sampled \textit{KDo} uniform randomly $20\%$ of the time. We performed multiple runs
and the results in Fig. \ref{fig:extrasim} show that including all the relevant variables gives a better model. More the number of missing variables, more the number of experiments needed to identify them all. Moreover, it becomes progressively harder. As a variable gets included in the model, possible \textit{Situation}s increase.
While evaluating $\text{R}(o,\text{A}_{\text{j}}|\text{S})$ (Eq. (\ref{coeffmutualInf})), we only consider the distributions of $o$ conditioned over values of $\text{A}_{\text{j}}$ that have been observed more than a certain number of times in the \textit{Situation} S. 
The active learning algorithm avoids repeating \textit{Situation}s and thus it becomes incrementally harder to identify relevant variables.

\vspace{-0.3cm}
\subsection{Pickup Task}
\begin{figure}
\begin{subfigure}{.48\textwidth}
\begin{tikzpicture}
[->,>=stealth',shorten >=1pt,node distance=1.8cm, minimum size=1.2cm, main node/.style={circle,draw},additional node/.style={circle,dashed,draw}]
  \node[main node] (Size) {\small{\textit{Size}}};
  \node[main node] (Pick) [below of=Size] {\small{\textit{Pick}}};
  \node[main node] (Shape) [left of=Size] {\small{\textit{Shape}}};
  \node[main node] (Weight) [right of=Size] {\small{\textit{Weight}}};
  \node[main node] (Arm) [right of=Pick] {\small{\textit{Arm}}};
  \path[every node/.style={font=\sffamily\small}]
    (Size) edge [right] node[left] {} (Pick)
    (Weight) edge node [right] {} (Pick)
    (Shape) edge [right] node {} (Pick)
    (Arm) edge node [right] {} (Pick);
\end{tikzpicture}
\caption{}
\end{subfigure}  
\begin{subfigure}{.48\textwidth}
    \null
    \begin{algo}
    SIT = \{\textit{Shape}, \textit{Size}, \textit{Weight}, \textit{Arm}\}\\
    CNTX = \{\textit{Shape}, \textit{Size}, \textit{Weight}\},\\
   COMM = \{\textit{Arm}\}\\
    OUT = \{\textit{Pick}\}, ATT$_{\text{\textit{Pick}}}$ = \{\}
\vspace{-0.2cm}

\noindent\hrulefill 
\vspace{0.15cm}
\caption{}
\label{mpickupset}
\vspace{0.2cm}
\noindent\hrulefill 

\vspace{0.02cm}
   \textit{Arm} $\epsilon$ \{\textit{Left},\textit{Right}\},\\
    \textit{Shape} $\in$ \{\textit{Ball},\textit{Box},\textit{Cylinder}\},\\
    \textit{Size} $\in$ \{\textit{Small},\textit{Large}\},\\
    \textit{Weight} $\in$ \{\textit{Light},\textit{Heavy}\},\\
    \textit{Pick} $\in$ \{\textit{Success},\textit{Failure}\},
 \end{algo}
 \vspace{-0.3cm}
\caption{}
\label{mpickupvars}
\end{subfigure}
\vspace{-0.1cm}
\caption{Capability Model for the \textit{Pickup} task. (a) depicts the Bayesian Network, (b) shows the type of each variable in the model and, (c) describes the values each variable can take.}
\label{mpickup}
\end{figure}
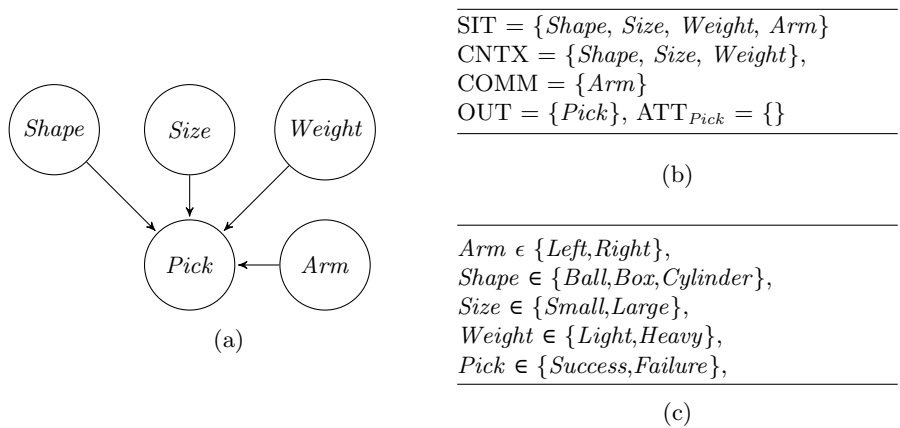

\vspace{-0.1cm}
We programmed a Pepper robot to detect and pickup objects of three shapes viz. spherical, cubical and cylindrical. We experimented with two sets of weights and two sizes for each shape (in total 12 types of objects). Fig. \ref{mpickup} depicts the Capability Model for the \textit{Pickup} task.
A \textit{Context} (\textit{Size}, \textit{Shape} and \textit{Weight}) denotes an object type. In every experiment Pepper was asked to pick up a particular type of object with one of its arms.

We performed 2 trials with 70 experiments each to learn the conditional probability tables associated with the Capability Model. We computed  $\text{\textit{Score}}_{\text{\textit{Pickup}}}$, as defined in Eq. (\ref{eq:contextcapab}), for each object type using the reference $P^{\text{\textit{Pickup}}}_{\text{ref}}$ defined in Eq. (\ref{eq:pexpectpickup}). We identified object types with score higher than a threshold at the end of both trials, as favourable for Pepper to pickup.
\begin{equation}
\label{eq:pexpectpickup}
    P^{\text{\textit{Pickup}}}_{\text{ref}}(\text{\textit{Pick}}|\text{\textit{Arm}}
    )=
    \left\{
    \begin{array}{ll}
    1& \text{if \textit{Pick} = \textit{Success}}\\
    0 & \text{otherwise} \\
    \end{array}
    \right.    
\end{equation}
Having knowledge of the scenarios a robot is well suited for could help in a collaborative task. To demonstrate this, we employed the robot along with a human to clear a cluttered table. In every experiment, the human cleared all but 4 objects off the table and, the robot had to clear the rest. The robot was allowed 3 tries per object (12 in total).
We performed such experiments in two settings. In the first setting, the human randomly selected objects for the robot to pick up and in the second, the human only selected objects of favourable types. We conducted 5 experiments in each setting.
Table \ref{clearthetableres} summarizes the results.  
\label{subsec:mapplicable}
\begin{center}
    \begin{tabular}{ c | c | c }
    Settings & Number of objects cleared & Number of tries\\
    \hlineB{2.5}
    Objects of any type & 1.8 $\pm$ 0.98 & 8.8 $\pm$ 1.7\\
    \hline
    Objects of favourable types & 3.4 $\pm$ 0.49 & 6.6 $\pm$ 1.2\\
    \end{tabular}
    \captionof{table}{Results of the clear-the-table task ($\mu\pm\sigma$) after 5 experiments per setting.
    } 
    \label{clearthetableres}
\end{center}
Performance at the task is better in terms of number of tries as well as number of objects cleared, when the robot is employed in a favourable scenario.

A video outlining our work can be found at \url{https://youtu.be/_9fm3U80vHE}.
\vspace{-0.4cm}
\vspace{-1.1cm}
\section{Conclusion and Future Work}
\vspace{-0.2cm}
Building models of unknown robots becomes all the more relevant as robots become increasingly multifunctional. To the best of our knowledge, the problem of inferring capabilities from appearance and active experimentation is novel and yet unexplored. We presented an algorithm to build capability models from experiments and showed results with a NAO and a Pepper robot at two different tasks. Experimenting with a physical system requires an operator. However, we still need a systematic approach to design experiments. Our algorithm can serve as a tool for humans in determining, "What can this robot do?" 

As an extension to this work we intend to develop an approach to draw inferences about the capabilities of a robot from its appearance and specifications. A robot's capabilities may have certain limitations which are intrinsic and others which it may learn to overcome. Enabling the learner to distinguish between the two and experiment accordingly is an interesting future direction. Moreover, we assumed that experimenting with the subject in any \textit{Situation} has the same cost. Depending upon the task this may be quite far from reality. Learning models in such scenarios is another interesting problem.  
\vspace{-0.4cm}
\section{Acknowledgements}
\vspace{-0.1cm}
This research is partially sponsored by DARPA under agreements
FA87501620042 and FA87501720152 and NSF under grant IIS1637927. 
The views and conclusions contained in this document are those of the authors only.
\vspace{-0.4cm}
\bibliographystyle{IEEEtran}
\bibliography{IEEEabrv,imports/bibfile}

\end{document}